\documentclass[letterpaper, 10 pt, conference]{ieeeconf}

\IEEEoverridecommandlockouts                              
\usepackage{amsmath} 
\usepackage{amssymb}  
\usepackage{cite}

\let\labelindent\relax
\usepackage{enumitem}

\usepackage{bbm}
\usepackage{algorithm}

\usepackage{algorithmic}

\def\RR{\mathbb{R}}

\def\AA{\mathcal{A}}

\def\TT{\mathcal{T}}
\def\DD{\mathcal{D}}
\def\SS{\mathcal{S}}

\newlist{HL}{enumerate}{1}
\setlist[HL]{label=\textbf{H\arabic*:}}

\newlist{ML}{enumerate}{1}
\setlist[ML]{label=\textbf{M\arabic*:}}

\usepackage{tikz}
\usepackage{amsmath}
\usetikzlibrary{matrix}
\usepackage{float}

\DeclareMathOperator*{\argmax}{arg\,max}

\usepackage[font=footnotesize]{caption}

\vspace{-0.25in}

\title{\LARGE \bf
Coordination With Humans Via Strategy Matching
}

\author{Michelle Zhao, Reid Simmons, Henny Admoni \\
\thanks{Authors are with the Robotics Institute, Carnegie Mellon University, Pittsburgh, PA, USA
{\tt\small \{mzhao2, rsimmons, hadmoni\}@andrew.cmu.org}}}
\begin{document}

\maketitle

\begin{abstract}
Human and robot partners increasingly need to work together to perform tasks as a team. Robots designed for such collaboration must reason about how their task-completion strategies interplay with the behavior and skills of their human team members as they coordinate on achieving joint goals. Our goal in this work is to develop a computational framework for robot adaptation to human partners in human-robot team collaborations. We first present an algorithm for autonomously recognizing available task-completion strategies by observing human-human teams performing a collaborative task. By transforming team actions into low dimensional representations using hidden Markov models, we can identify strategies without prior knowledge. Robot policies are learned on each of the identified strategies to construct a Mixture-of-Experts model that adapts to the task strategies of unseen human partners. We evaluate our model on a collaborative cooking task using an Overcooked simulator. Results of an online user study with 125 participants demonstrate that our framework improves the task performance and collaborative fluency of human-agent teams, as compared to state of the art reinforcement learning methods.
\end{abstract}
\vspace{-0.07in}

\section{Introduction}
\label{sec:introduction}
Robots increasingly serve as collaborative partners in applications where humans cannot operate alone, such as robot-assisted elder care \cite{mertens2011assistive} and cooking \cite{koppula2016anticipatory}. In such applications, humans may employ any number of equally reasonable \textit{strategies} to achieve their goals. Robots designed for collaboration should be able to adapt their behavior in order to coordinate with the different strategies employed by different human partners \cite{cooperation}.

Poor coordination between partners can lead to inefficient collaboration on tasks. Consider a robot and human partner working together to make a sandwich. Both partners get condiments, believing the other is getting the bread. This disfluency results in task inefficiencies: production time increases since one agent now must locate the bread before they are able to begin assembling the sandwich; and cleanup time increases since double the amount of condiments have been taken out. In collaborative tasks without explicit verbal communication, teams can be even more susceptible to disfluencies like these. In order to overcome disfluencies, team members must be able to coordinate by inferring their partner's strategy from their observable actions. 

In this work, we explore the following question: \textit{How can a robot partner recognize the task strategy employed by a human partner, and adapt its own response online?}
Prior work leverages human-human team demonstrations to learn robot behavior policies for collaborative tasks \cite{carroll2019utility}, using human data in aggregate, without separating the demonstrations by strategy. 
When a dataset contains one class that is underrepresented, the trained model often generalizes well to the majority class but poorly to the minority class, a problem addressed by techniques including undersampling \cite{undersampling}. This type of aggregate model performs well with the ``average'' or most common human behavior, but may miss underrepresented strategies. On the other hand, an agent that can distinguish between strategies will likely generalize better to a greater diversity of human behavior. 

We propose a method for a robot to identify and adapt to discrete, task-oriented strategies that determine the team's behavioral patterns. We situate our work in the simulated Overcooked domain, a human-robot collaborative cooking testbed. Using data collected in \cite{carroll2019utility} of human-human team task demonstrations for five different kitchen environments, we annotate the trajectories to represent them as high level task sequences. Next, we transform the annotated sequences of team actions into low dimensional representations using hidden Markov models. Clustering on the low-dimensional representations extracts groups of similar team behavior, which define discrete strategies representing different team approaches employed on a collaborative task. Robot policies are trained using apprenticeship learning \cite{apprenticeship} to imitate distinct strategies. The resulting agent is a Mixture-of-Experts model that maintains a dynamic belief over the strategy space for unseen human partners at test time. 

We conducted an online user study to investigate the utility of coordination on behavior in human-robot teams. 125 participants performed a collaborative task with our proposed agent as well as with an existing baseline agent \cite{carroll2019utility} in the five environments for which we had data. Our approach improved team task performance in two of the five tested environments and team collaborative fluency in three of the five tested environments. 

\begin{figure}[t]
    \centering
    \includegraphics[width=0.49\textwidth]{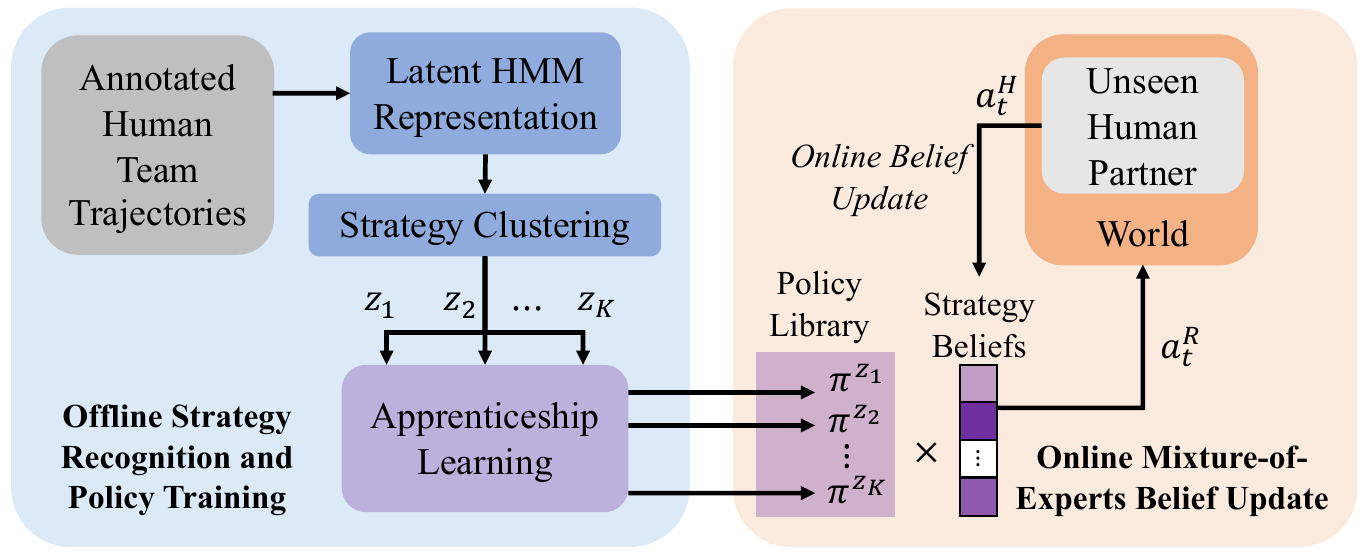}
    \caption{The offline training paradigm consists of strategy recognition and training a policy library of strategy-specific agent policies. Beliefs over strategies are updated online during interactions with new partners. Actions taken by the robot are sampled from a belief-weighted combination over action distributions generated by each strategy-specific policy.}
    \label{fig:sys_arch}
    \vspace{-0.25in}
\end{figure}

\section{Related Work}
\label{sec:related_work}
\subsubsection{Ad-Hoc Teaming}
Ad-hoc teaming in Human Robot Interaction (HRI) requires the ability of robot agents to adapt to unseen partners \cite{noauthor_making_nodate, dafoe_open_2020}, who may differ in knowledge, skill, and behavior. Prior work \cite{noauthor_making_nodate} proposes a general purpose algorithm that reuses knowledge learned from previous teammates or experts to quickly adapt to new teammates. The approach takes two forms: (1) model-based, which develops a model of previous teammates' behaviors to predict and plan in online, and (2) policy-based, which learns policies for previous teammates and selects an appropriate policy online. Another important challenge in ad-hoc teaming is modeling uncertainty over partner characteristics \cite{agmon_modeling_nodate, stone_ad_2010}. In the Overcooked environment, \cite{carroll2019utility} showed that incorporating human models learned from data improves the performance of agents compared to agents trained to play with themselves. Instead of training agents to partner with a general human proxy model as in \cite{carroll2019utility}, we train a library of strategy-specific agent policies that represent different coordination behavior patterns. Distinguishing strategy allows for a policy library that captures differences in team coordination patterns that may otherwise wash out in a single general model.

\subsubsection{Multi-agent Reinforcement Learning}
In cooperative multi-agent settings, self-play (SP) trains a team of agents that work well together. A collaborative agent that excels with the partners with which it was trained may not generalize well to new partners at test time, especially when the new partners differ significantly from the pool used for training \cite{cui_k-level_nodate}. Other-play (OP) \cite{hu_other-play_2021} addresses this problem, demonstrating improved zero-shot coordination with human-AI performance on the Hanabi game \cite{hanabi}. A self-play training paradigm that assembles agents representing untrained, partially trained, and fully trained partners by extracting agent models at different checkpoints in the training duration has been shown to produce robust agents trained on the suite of partners \cite{strouse_collaborating_2022}. Prior work \cite{he2016opponent} models opponents in deep multi-agent reinforcement learning settings by training neural-based models on the hidden state observations of opponents. A Mixture-of-Experts architecture maintains a distribution over different opponent strategies, allowing this model to integrate different strategy patterns.

\subsubsection{Adaptation in Human-Robot Interaction}
Past research has studied how robots can adapt and learn from human partners. Key to robot-to-human adaptation is understanding people's behavior through observation. Markov Decision Processes (MDPs) are a common framework for goal recognition \cite{ngirl}. By learning a model of human intent and preferences \cite{intentionmotionplan}, robots can reason over different types of human partners \cite{collabaction,Nikolaidis2013HumanrobotCC}. Similar in vein to our work, \cite{indmutual} applied a best-response approach to selecting policies from a library of response policies that best match a particular player type. Building an understanding of the human partner requires multi-faceted models of humans that capture nuanced differences. Our work on adaptation focuses primarily on adapting robot behavior to the task approach (strategy) of a human partner. Our adaptation approach is similar to \cite{nikolaidis2015efficient}, where human demonstrations are clustered into dominant types and a reward function is learned for each type, for which Bayesian inference is used to adapt to new users.

\vspace{-0.05in}

\section{Preliminaries}
\label{sec:preliminaries}

\subsubsection{Task Scenario}
\label{sec:task_scenario}
\begin{figure}[t]
    \centering
    \includegraphics[width=0.45\textwidth]{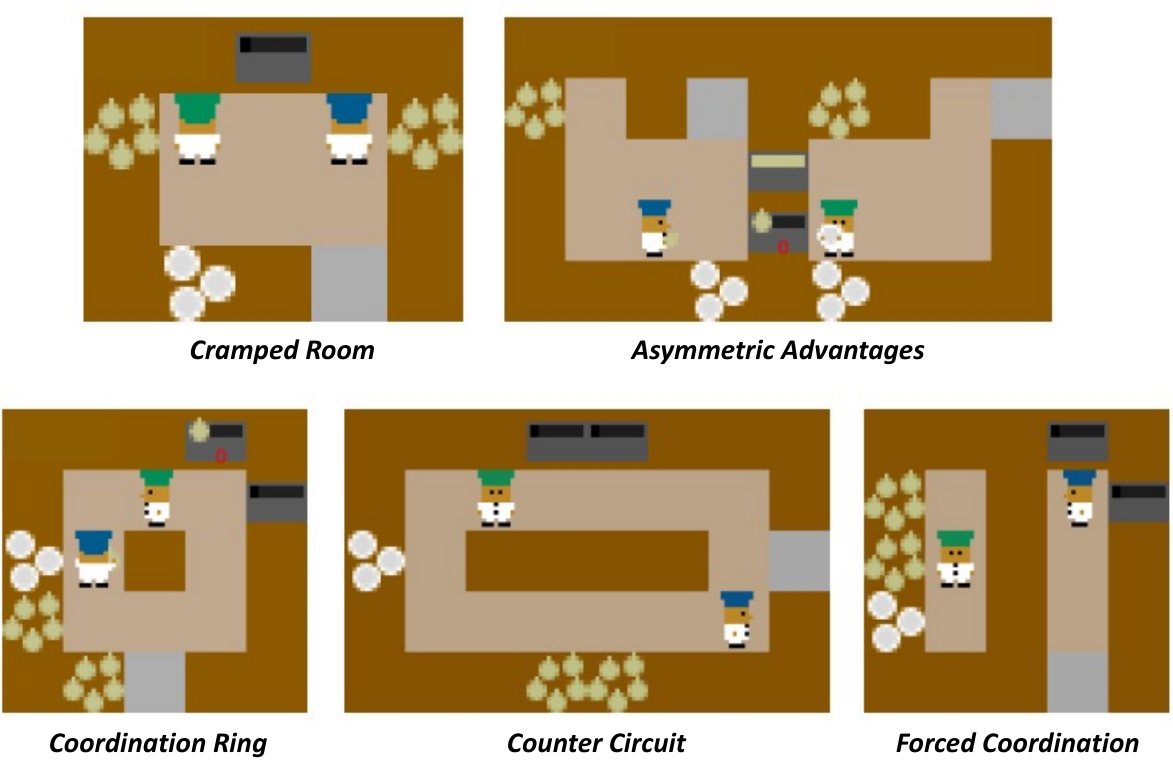}
     \setlength{\belowcaptionskip}{-20pt} 
    \caption{The Overcooked experimental layouts. Environments vary in the amount of constrained space, actions available to different player positions, and interdependence of player actions to achieve the objective.}
    \label{fig:overcooked_description}
    \vspace{-0.01in}
\end{figure}

In order to study human-robot collaboration, we study the \textit{Overcooked} environment \cite{carroll2019utility}, a collaborative cooking task. Dyads (consisting of robot agents or humans) collaborate in a constrained shared environment (Fig \ref{fig:overcooked_description}). Their objective is to prepare an order (onion soup) and serve it as many times as possible in an allotted time. 

\subsubsection{Strategies}
In the Overcooked task, agents must perform sequences of high-level tasks to serve orders. Examples of high-level tasks include picking up onions and plates, placing onions into pots, and serving soup. Each high-level task requires a sequence of lower level subtasks (i.e. motion primitives). Teams collaborate on shared tasks in  different ways. For example, in role specialization, players take sole responsibility for particular tasks, whereas in complete-as-needed approaches, each partner performs the next required task. In addition to role-oriented strategies, collaborative approaches also prescribe the order in which tasks are performed. Teams that serve dishes while the next orders are cooking employ more time-efficient strategies. We define collaborative \textit{strategies} as the sequence in which high-level tasks are interleaved and distributed across teammates. Since actions of all team members are involved in task approach, strategy is computed at the team level.

\subsubsection{MDP Formulation}
The task is modeled as a two-player Markov decision process (MDP) defined by tuple $\langle \SS, \AA = \{ \AA^1, \AA^2\}, \TT, R \rangle$. $\SS$ is the set of states. The action space of a game with two players is $\AA = \AA^1 \times \AA^2$. The set of actions available to each player $i$ is $\AA^i$. The transition function $\TT$ determines how the state changes based on a joint action by both players, $\TT: \SS \times (\AA^1, \AA^2) \rightarrow \SS$. $R: \SS \rightarrow \RR$ is the team reward function. $\pi^i$ represents agent $i$'s policy. $Z = \{z_1, ..., z_K\}$ represents the set of possible team collaborative strategies. We further denote a policy that corresponds to strategy $z_k$ as $\pi_{k}$.

\section{Approach}
\label{sec:approach}
We introduce MESH (Matching Emergent Strategies to Humans) as an approach for coordination of collaborative strategies with human partners. The MESH approach is comprised of two components: strategy recognition and Mixture-of-Experts adaptation (Fig \ref{fig:sys_arch}). In the strategy recognition step, MESH learns the latent strategies employed by different human teams using unsupervised learning on team trajectories. Since strategies are learned over teams, the MESH agent aims to coordinate on the same strategy employed by the human partner. Once team collaborative strategies are identified, we apply apprenticeship learning to train policies that imitate each extracted strategy. These strategy-specific policies are combined to form a Mixture-of-Experts model \cite{moe_survey}. When interacting with unseen human partners at test time, the Mixture-of-Experts model adapts its belief over strategies to the inferred strategy of the human partner.

\subsection{Strategy Recognition}
\label{sec:unsupervised}
The collaborative strategy of a team is represented as a single learned parameter, which determines how teams order and interweave tasks between members. The robot's objective is to use observations to identify what strategies exist that human teams may employ. We utilize data collected in \cite{carroll2019utility}, containing observed trajectories of human-human dyads in five kitchen environments, illustrated in Figure \ref{fig:overcooked_description}.

We frame identifying collaborative strategies from observations of teams as an unsupervised learning task of recognizing patterns in the sequences of high-level tasks performed by both players. The action space $\AA^i$ consists of 5 actions: \{Move $\times$[North, South, West, East], Interact\}, where the Interact action encompasses all high-level actions where the agent interacts with the environment by picking up or placing objects. We translate environment interactions into a set of 7 high-level actions (subtasks) $G = \{g_1, .., g_m\} = $\{Pick-up $\times$[Onion, Dish, Soup], Place $\times$[Onion, Dish, Soup], Serve Soup\}. Using these subtask definitions, we construct an annotated dataset $\DD = \{\xi_1, .., \xi_M\}$ of $M$ team subtask trajectories. The sequence of high-level actions performed by both players defines a team $i$'s trajectory: $\xi_i = \{g^1_1, g^2_1, ...,g^1_t, g^2_t, .., g^1_T, g^2_T\}_i$, where $g^i_t$ represents the subtask performed by player $i$ at time $t$.

Since there are numerous subtask orderings that can define possible team coordination strategies, to maintain a tractable space of strategies to reason over, only orderings that differ significantly should be classified as distinct strategies. MESH learns a low-dimensional sequence representation of team action sequences through a Hidden Markov Model (HMM). HMMs are a formulation for learning probabilistic models of linear sequences that allow a probability for a sequence of observable events to be computed. The HMM is specified by the following components:
\begin{enumerate}
    \item The set of observations to the HMM is $O = \{g^1_1, g^2_1, ...,g^1_t, g^2_t, .., g^1_T, g^2_T\}_{\xi_i \in \DD}$, where the sequence of observations is the sequence of high-level tasks performed by both players. 
    \item $X = \{x_1, x_2, .., x_N\}$ represents a set of $N$ hidden states.
    \item $B = X \times X$ is a transition probability from hidden state $x_i$ to hidden state $x_j$.
    \item $C = X \times O$ is an emission probability from hidden state $x_i$ to observation $g^i_t$.
\end{enumerate}

The HMM takes as input the high-level subtask sequences of each team in the dataset. Given these observations, the Baum-Welch algorithm \cite{baumwelch}, an expectation-maximization approach to learning HMM parameters, is used to learn the transition matrices $B$ and $C$. We next apply the Viterbi algorithm \cite{viterbi} to compute the most likely hidden state sequence for each sequence of observations, representing an underlying temporal structure in the team action sequences. Since the number of hidden states $N$ is chosen to be less than $|G|$, the hidden state sequence provides a low-dimensional representation, $ \bar{x}_i = (x_0, x_1, ..., x_N)_{\xi_i}$ $\forall \xi_i \in \DD$ of each team's observed sequence of subtasks. The sequences $\bar{x}_i$ are trimmed to length.

\begin{figure*}
  \begin{minipage}{.99\textwidth}
  \centering
    \includegraphics[width=0.99\textwidth]{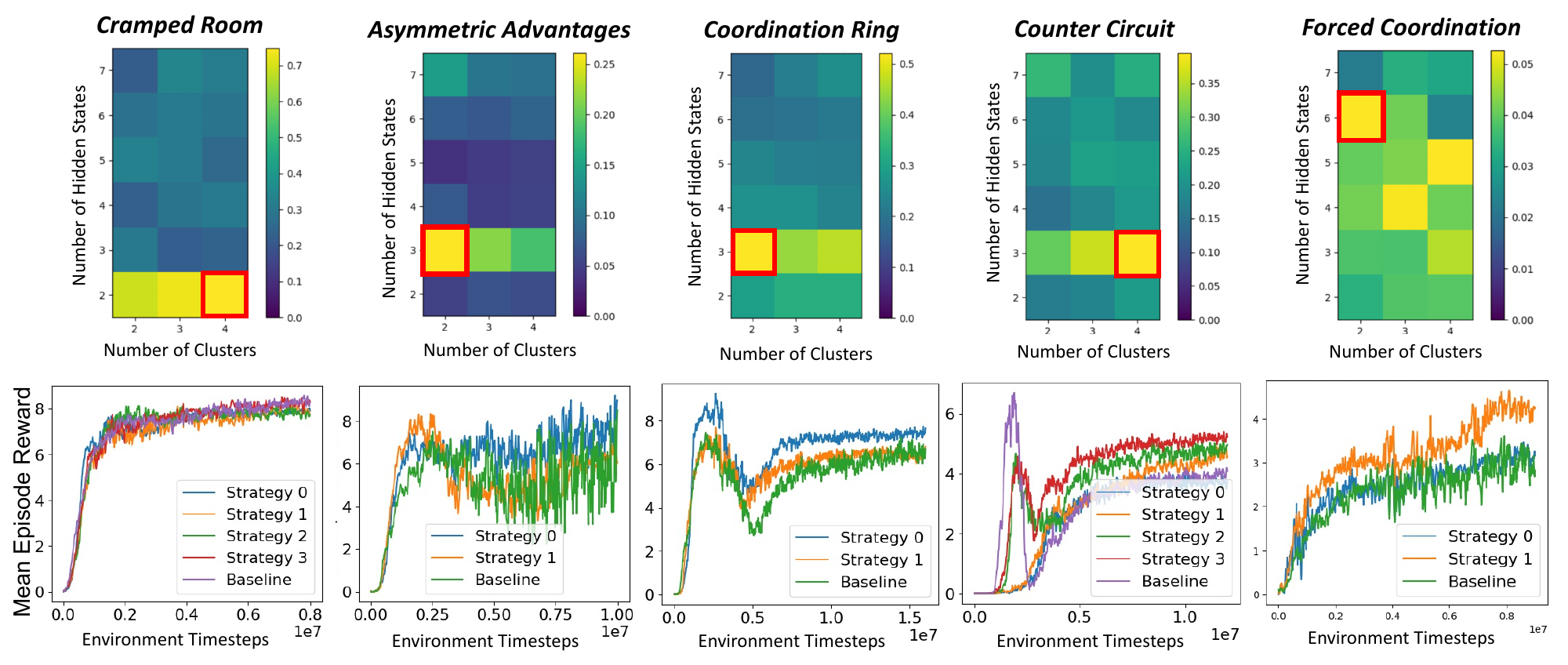}
     \setlength{\belowcaptionskip}{-20pt} 
    \caption{(Top Row) The strategy clusters for each environment are selected by taking the model achieving the highest silhouette score. Silhouette score is more correlated with the number of hidden states than number of clusters. (Bottom Row) Training rewards for each strategy-specific policy and the baseline. Strategy specific policies achieve varying rewards in training. The same training parameters were used as in the baseline \cite{carroll2019utility} for fair comparison.}
    \label{fig:silscore}
  \end{minipage} \quad
\end{figure*}

The discrete strategy $z$ for each team is learned by clustering over the low-dimensional sequence representation. K-means clustering over the set of hidden-state sequences, $\bar{x}_i$ $\forall \xi_i \in \DD$, identifies teams that share the same strategy. The final clusters define strategy classifications $z \in Z$. We performed grid search over the number of strategy clusters $K$ and number of hidden states $N$, measuring the silhouette score of each combination. Silhouette score \cite{silhouette} evaluates clustering algorithms by comparing how similar an element is to its own cluster (cohesion) compared to other clusters (separation). We selected the combination of number of hidden states and clusters achieving the highest silhouette score to determine the strategy classification for each environment (Figure \ref{fig:silscore}).

\begin{figure}[t]
\begin{algorithm}[H]
\caption{Apprenticeship Learning of Strategy-Specific Policies}
\label{algo:apprenticeship}
\begin{algorithmic}[1]
\FOR{Strategy $k=1,\;\cdots,\;K$}
\STATE $\DD_k = \{\xi_i : \text{team } i \in \text{strategy cluster }z_k\}$ \hfill 
\STATE $\pmb{\theta}_k \leftarrow \textsc{MaxEnt IRL}(\DD_k)$ \hfill 
\STATE $f^h_k \leftarrow \textsc{Behavior Cloning}(\DD_k)$ \COMMENT{Train behavior cloning human model}\hfill 
\STATE $\pi_k \leftarrow \textsc{PPO}(\pmb{\theta}_k, f^h_k)$ \COMMENT{Compute agent policy using PPO with BC human model partner.}\hfill 
\ENDFOR
\STATE \textsc{Strategy-specific Policy Library} = $\{\pi_1, .., \pi_K\}$
\end{algorithmic}
\end{algorithm}
\vspace{-0.45in}
\end{figure}

\subsection{Mixture-of-Experts Model}

\subsubsection{Apprenticeship Learning for Training Strategy-specific Agents}
\label{sec:partners}

Apprenticeship learning (AL) \cite{abbeel_apprenticeship_2004} is the task of learning from expert demonstrations. In order to train agents that employ particular strategies $z\in Z$, we employ the apprenticeship learning algorithm outlined in \cite{abbeel_apprenticeship_2004}, combined with the Overcooked agent training approach introduced in \cite{carroll2019utility} (Algorithm \ref{algo:apprenticeship}). For each collaborative strategy $z_k \in Z$, we train a human proxy model $f^h_k$ to represent a partner employing strategy $z_k$. The human proxy model $f^h_k$ is trained using behavioral cloning (BC) on all team trajectories in aggregate with fine-tuning on the subset of trajectories belonging to teams from the same strategy cluster $z_k$. Next, we learn the reward weights $\theta_k$ corresponding to strategy $z_k$ using Maximum Entropy Inverse Reinforcement Learning \cite{maxent}. The reward features are: $\phi=$ \textit{\{onion placed in empty pot, onion placed in partially-filled pot, dish picked up, soup picked up from pot, both pots full, soup served\}}. We construct a policy library of strategy-specific agents, trained to employ only their corresponding strategy. Using Proximal Policy Optimization (PPO) \cite{ppo}, we compute the strategy-specific optimal policy $\pi_k$ for the MDP using rewards $R = \theta_k^T \phi$. Under a similar training regime to \cite{carroll2019utility}, the policy $\pi_k$ is trained with BC partner $f^h_k$.

\subsubsection{Adaptation of Human Partner Strategies}
Leveraging the notion of best response from game theory \cite{gametheory_textbook}, the adaptation task of the robot at test time is to match its actions to a new partner's strategy. The MESH agent uses a discounted-memory Mixure-of-Experts (MOE) model \cite{moe_survey} to coordinate with the inferred strategy of the partner. The strategy of the human partner is predicted based on how well each of the candidate strategies in the policy library predict the human's behavior, with an emphasis on recent human actions. The agent takes a weighted combination over the action probability distributions proposed by each strategy-specific policy. It selects the action with the highest probability after the weighted voting. The next paragraph details the mixture approach (Algorithm \ref{algo:adapt}). 

The strategy-specific robot policy library is $\{\pi_1, .., \pi_K\}$, where $K$ is the number of strategies identified in the given environment. $\pi^R_k$ indicates the policy employing strategy $z_k$ at the position of the robot player. $\pi^H_k$ represents using the strategy-specific robot policy $\pi_k$, but taken from the position of the human player. The policies map states to a distribution over actions, which are normalized to form a probability distribution. $\pi^R_k(a_i|s_t)$ denotes a scalar valued probability of action $a_i$ being taken by the robot in state $s_t$.

\begin{figure}[t]
\begin{algorithm}[H]
\caption{Online Strategy-Adaptive Mixture-of-Experts}
\label{algo:adapt}
\begin{algorithmic}[1]
\STATE \textsc{Strategy Policy Library} = $\{\pi_1, .., \pi_K\}$ \hfill
\STATE \text{strategies} = $\{z_1, .., z_K\} \in Z$ \hfill
\STATE $\mathbf{w}_0 = [\frac{1}{K}, .., \frac{1}{K}]$ \hfill
\FOR{$t=1,\;\cdots,\;T$}
\STATE $a_t^R = \argmax_{a_j\in \AA} \left[ \sum_{k=1}^K \pi^R_k (a_j|s_t) w_t^k \right]$ \hfill
\STATE Human partner takes action $a_t^H$ \hfill
\STATE $d_t^k = \sum\limits^t_{j=1}\gamma^{t-j} \max\limits_{a_i \in \AA} \pi^H_k(a_i|s_{t-j}) -  \pi^H_k(a_t^H|s_{t-j})$ \hfill
\STATE $\tilde{d}_t^k = \frac{d_t^k}{\sum\limits_k d_t^k }$ \hfill
\STATE $ w_{t+1}^k = \frac{1-\tilde{d}_t^k}{\sum\limits_k 1-\tilde{d}_t^k} $ \hfill

\ENDFOR
\end{algorithmic}
\end{algorithm}
\vspace{-0.45in}
\end{figure}

The strategy-specific policies are weighted by comparing each policy's predicted action to the true action taken by the human. When the highest probability action predicted by the strategy-specific policy is the action that the human took, this difference is 0, since $\max_{a_i \in \AA} \pi^H_k(a_i|s_{t-j})] =  \pi^H_k(a_t^H|s_{t-j})$. Thus, $d_t^k = \sum^t_{j=1}\gamma^j \max_{a_i \in \AA} \pi^H_k(a_i|s_{t-j}) -  \pi^H_k(a_t^H|s_{t-j})$ is a measure of how inaccurately the robot's human model represents the human's actions. The higher $d_t^k$, the worse the human model. $\tilde{d}_t^k$ is the normalized value of $d_t^k$ over strategies. The belief weights over strategies, $\mathbf{w}^t$, are the normalized, inverted $\tilde{d}_t^k$ values. The belief model sees only the actions taken by the human partner and does not have access to a true human model. Assuming that the human is sampling from some unknown true human distribution, the expected action taken by the human is maximum likelihood action under their true, but unknown, policy. Because of this limited fidelity feedback from the human, we evaluate the maximum probability predicted human actions of each strategy policy. The $\gamma$ factor represents a discounted memory factor, which prioritizes recent human actions, allowing the likely strategy of a player to shift as the episode progresses.

\section{Simulation Results}
\label{sec:simulation}

\begin{figure}[t]
    \centering
    \includegraphics[width=0.49\textwidth]{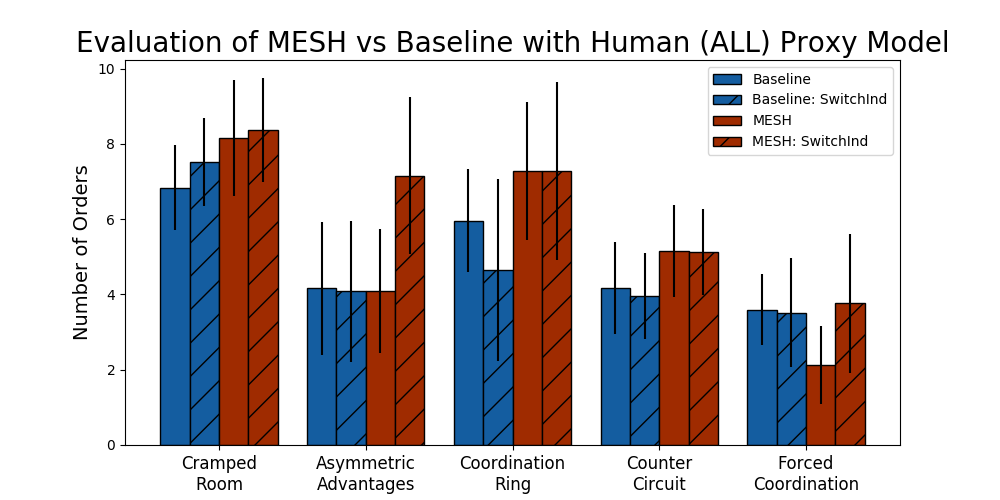}
    \caption{The MESH agent generally outperforms the Baseline model while collaborating with the human-(ALL) proxy model, except for in one position in the Forced Coordination and Asymmetric Advantage environments. The number of orders is averaged over 25 simulated games. The error bars represent standard deviation.}
    \label{fig:training}
    \vspace{-0.25in}
\end{figure}

We first compare the MESH agent to the baseline by evaluating the performance of both agents with a human proxy model trained using behavioral-cloning over all teams in the dataset. We refer this model the human-(ALL) proxy model. Switched indices (SwitchInd) refers to swapping the agent and human player positions. The MESH agent outperforms the Baseline agent in both player positions for four out of five environments. In Forced Coordination, the MESH agent outperforms on one position, while performing worse relative to the Baseline in the other position (Fig \ref{fig:training}).

\section{User Study}
\label{sec:experimental_design}

We ran a user study to compare collaboration of human partners with the MESH agent and a baseline agent that was trained using PPO with a single behaviorally-cloned human partner \cite{carroll2019utility}. Each participant performed a collaborative cooking task with both the MESH agent, and the Baseline agent. The study was a $5\times 2$ within-subjects design. The first repeated variable was (1) Kitchen Environment, of which there were 5 levels (layouts: [Cramped Room, Coordination Ring, Asymmetric Advantages, Counter Circuit, Forced Coordination]). The second repeated manipulated variable was (2) agent adaptability, which consisted of two types: MESH agent (adaptive), and (2) Baseline agent trained with PPO with a behaviorally-cloned human proxy (non-adaptive). The MESH Agent is adaptive since it is trained to imitate different strategies learned from observing human teams, allowing it to adapt its mixture policy to the strategy it believes the human partner is playing.

Participants were given a scaffolded training task to familiarize themselves with the online game's controls. Next, participants played all 5 environments of Overcooked with each of the two agents, for a total of 10 rounds. Each round is 60 seconds. The 5 environments were played sequentially, meaning that for each environment, participants will play with one type of agent, followed by the other, before moving onto the next environment. The order of environments and the order in which agents were presented was counterbalanced. We compared the two team compositions (human-MESH, human-Baseline) on task performance and collaborative fluency. Collaborative fluency (CF) \cite{fluency} is defined as the coordination of joint activities by members in a team. After each round, the participant rated the collaborative fluency of the team. Between environments, after playing with both agents, participants evaluated which agent they preferred.

\subsection{Hypotheses}
We hypothesize that participants collaborating with the MESH agent will demonstrate better task performance and higher collaborative fluency, measured through subjective and objective measures.

\begin{HL}
\item Participants will perform better on the cooking task by serving more dishes when paired with MESH.
\item Participants will prefer to team with MESH agents.
\item Teams with MESH agents will exhibit higher collaborative fluency, measured through objective measures of idle time, concurrent activity, and functional delay.
\item Teams with MESH agents will self-report higher subjective team collaborative fluency.
\end{HL}

\subsection{Measures}
As we seek to develop a more collaborative agent, collaborative fluency metrics measure how effective an agent is at coordinating its behavior with its human partner. While fluency measures do not always directly correlate with task efficiency, they can change people's perception of collaboration.

\subsubsection{\textbf{Task Performance}} Task performance is measured by the number of orders served by a team.
\subsubsection{\textbf{Preference}}  After each kitchen environment was played, participants were asked to select their preferred partner.
\subsubsection{\textbf{Objective Collaborative Fluency}}  Participant trials were analyzed and 3 objective fluency measures were computed (in seconds): (1) human idle time, (2) robot idle time, and (3) concurrent activity \cite{fluency}.
\subsubsection{\textbf{Subjective Collaborative Fluency}} Participants are asked 5-point Likert scale questions rating the team's fluency, fluency over time, the agent's ability to understand the actions of the participant, and the predictability of the agent's actions.
\begin{itemize}
    \item Partner X and I coordinated our actions well together.
    \item Partner X and I coordinated our actions better as the episode progressed.
    \item Partner X perceived accurately what tasks I was trying to accomplish.
    \item I was able to understand and predict what tasks Partner X was trying to accomplish.
\end{itemize}
We measured alignment between responses to these questions (Chronbach's $\alpha=0.92$ \cite{chronbach}), and the values were added into a single subjective fluency score.

\section{Results}
\label{sec:empirical_results}


We collected data from 148 Prolific \cite{prolific} workers over the age of 18. We removed 19 participants who did not complete the study or whose data was saved improperly and 4 participants who performed fewer than 10 keypresses in 1 or more trials (indicating deliberate lack of participation), resulting in 125 total participants. The keypress threshold was selected at 3 standard deviations away from the mean after skew adjustment with a square-root transform. We analyzed all numerical data using a two-way repeated measures ANOVA followed by post-hoc pairwise analyses with Bonferroni correction for multiple comparisons. Mauchly’s test of sphericity was performed on the data for each measure, and Greenhouse-Geisser correction was applied for the ANOVA if the sphericity assumption was not met. In the following graphs, $*$ denotes significance at $\alpha=0.05$, and $**$ denotes significance at $\alpha=0.01$. 58.4\% of the participants self-identified as female, 40.0\% as male, and 1.6\% as transgender, non-binary, or other. Participant ages ranged from 18 to 70 (M = 27.98, SD = 8.61). The recruitment procedure and study were approved by our Institutional Review Board. The study and hypotheses were pre-registered on Open Science Framework (10.17605/OSF.IO/JFW7T). 

\begin{figure}[b]
\vspace{-0.15in}
\center
\includegraphics[width=0.46\textwidth]{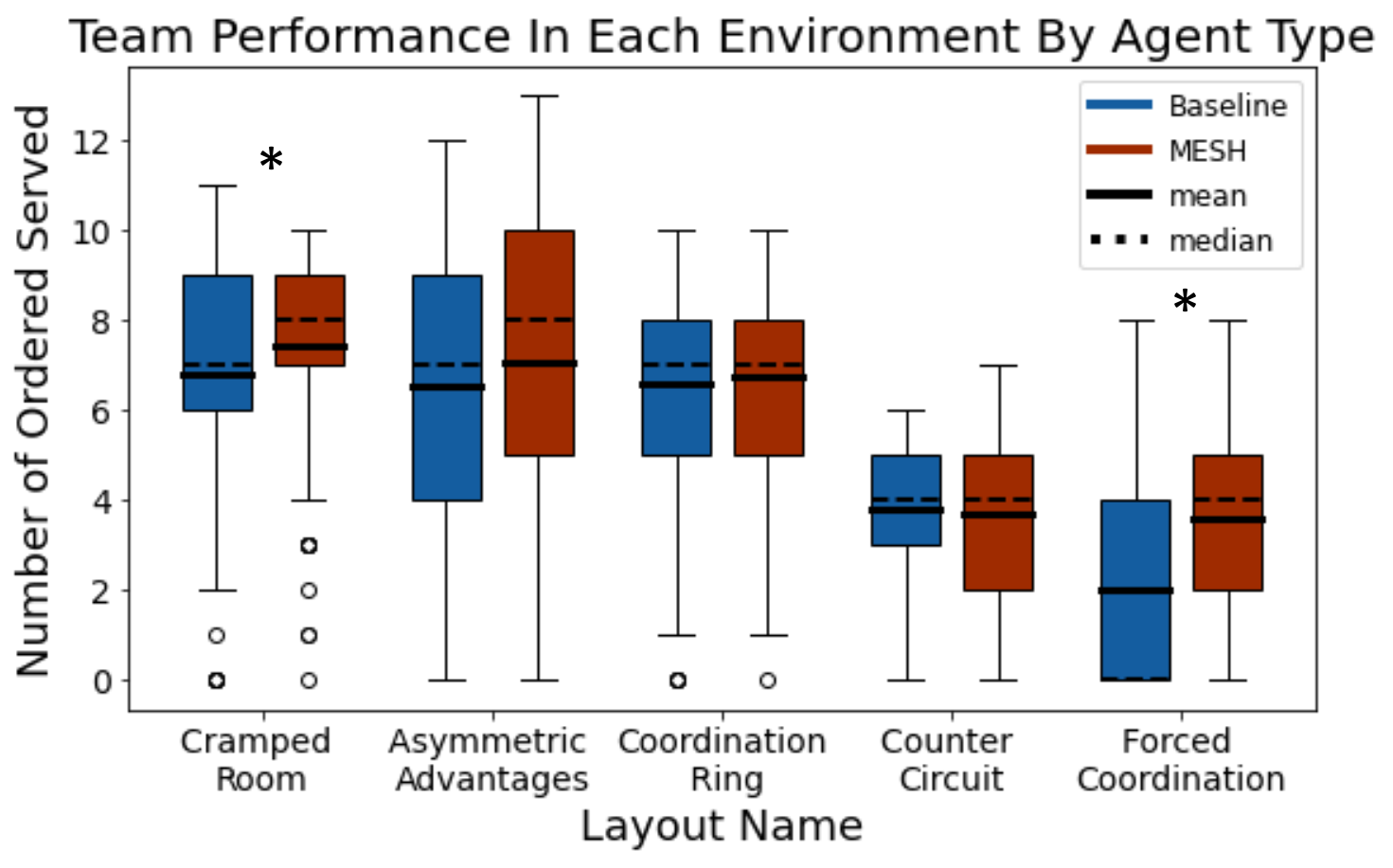}
\caption{Scores of human-MESH teams increased in Cramped Room and Forced Coordination environments. No significant difference was found in the other three environments.}
\label{score_boxplot}
\end{figure}

\subsubsection{Task Performance} \textbf{H1} is only supported for the Cramped Room and Forced Coordination environments (Fig \ref{score_boxplot}). In these layouts, participants served significantly more orders when paired with the MESH model over the Baseline model. For the other three environments, no significant difference between agents was observed. A two-way repeated measures ANOVA found that was a significant main effect of agent type on score ($F(1,124)=23.285, p<0.001, \eta^2 = 0.158$), where participants scored higher with the MESH agent over the baseline. There was a significant interaction between the layout and agent type, ($F(3.46,428.51)=8.275, p<0.001, \eta^2 = 0.063$). Post hoc testing with Bonferroni correction showed that for Cramped Room ($p=0.009$) and Forced Coordination ($p<0.001$), there was a significant increase in number of orders served with the MESH agent. 

\subsubsection{Preference} \textbf{H2} is only supported for the Forced Coordination environment. Agent preferences between the MESH and Baseline agents were queried via a forced choice preference question for each environment. A Chi-Squared Test measured the direction of participant preferences between agent types by comparing preferences against a uniform distribution. People preferred MESH for Forced Coordination: ($\chi^2=17.67, p < 0.001$). The baseline was preferred for Coordination Ring: ($\chi^2=8.712, p = 0.003$). The other environments did not exhibit significant differences.  

\begin{figure}[t]
\center
\includegraphics[width=0.46\textwidth]{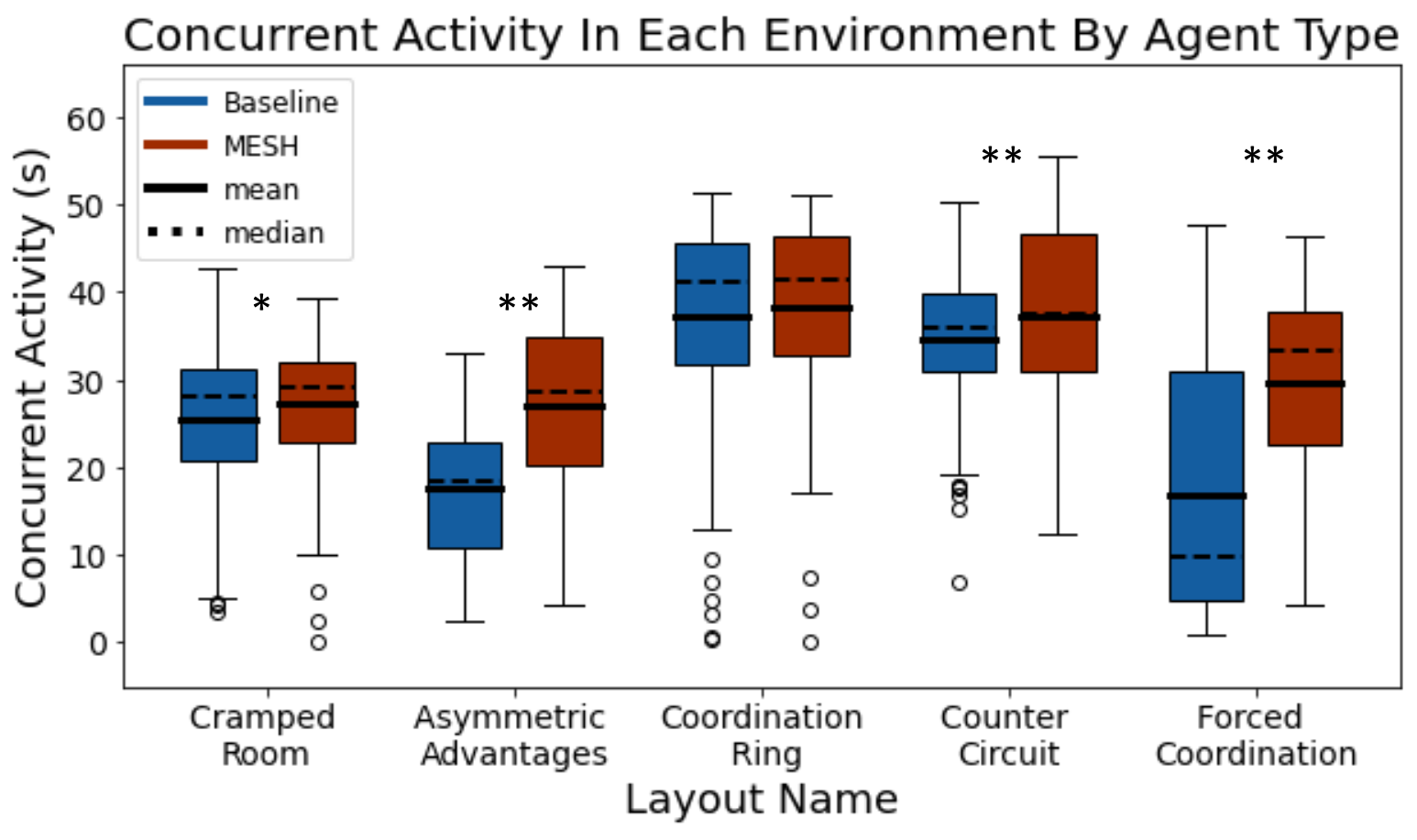}
\caption{Concurrent activity between human and robot partner was increased with MESH for four of five environments.}
\label{concurrent}
\vspace{-0.25in}
\end{figure}

\begin{figure}[b]
\center
\includegraphics[width=0.46\textwidth]{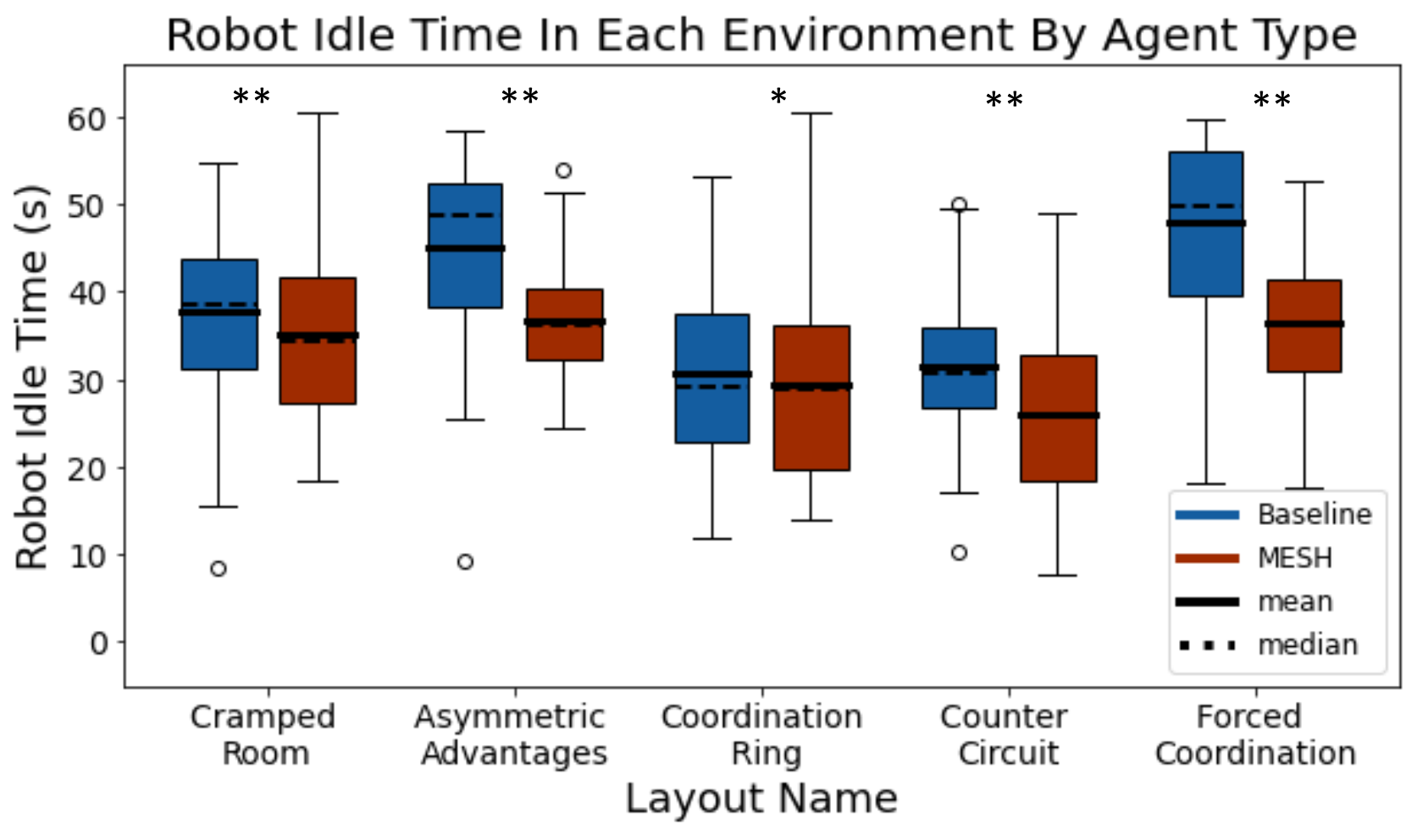}
\caption{Participants pairing with MESH agent saw decreased in robot idle time in all environments.}
\label{idle}
\end{figure}

\subsubsection{Objective Collaborative Fluency: Idle Time and Concurrent Activity} \textbf{H3} is supported in the majority of environments in concurrent activity and robot idle time. There was a significant main effect of agent type on the team concurrent activity ($F(1,124)=123.846, p<0.001, \eta^2 = 0.500$), where human-MESH teams displayed higher concurrent activity. There was a significant interaction between the layout and agent type on concurrent activity, ($F(2.611, 323.762)=27.983, p<0.001, \eta^2 = 0.184$). Post hoc testing with Bonferroni correction showed that for Cramped Room ($p=0.018$), Asymmetric Advantages ($p<0.001$), Counter Circuit ($p=0.004$), and Forced Coordination ($p<0.001$), there was a significant increase in concurrent activity with the MESH agent (Fig \ref{concurrent}). 

The results demonstrated a significant main effect of agent type on the robot idle time ($F(1,124)=144.604, p<0.001, \eta^2 = 0.538$), where human-MESH teams generally experienced lower robot idle time. There was a significant interaction between the layout and agent type on robot idle time, ($F(2.883, 357.535)=28.523, p<0.001, \eta^2 = 0.187$). For Cramped Room ($p<0.001$), Asymmetric Advantages ($p<0.001$), Coordination Ring ($p=0.047$), Counter Circuit ($p<0.001$), and Forced Coordination ($p<0.001$), robot idle time significantly decreased with the MESH agent (Fig \ref{idle}). 

We found a significant main effect of agent type on the human idle time ($F(1,124)=64.700, p<0.001, \eta^2 = 0.343$), where human-MESH teams experienced lower human idle time generally. There was no significant interaction between the layout and agent type on idle time ($F(2.714, 336.545)=2.538, p=0.063, \eta^2 = 0.020$).

\subsubsection{Subjective Collaborative Fluency.} 
\begin{figure}[t]
\center
\includegraphics[width=0.46\textwidth]{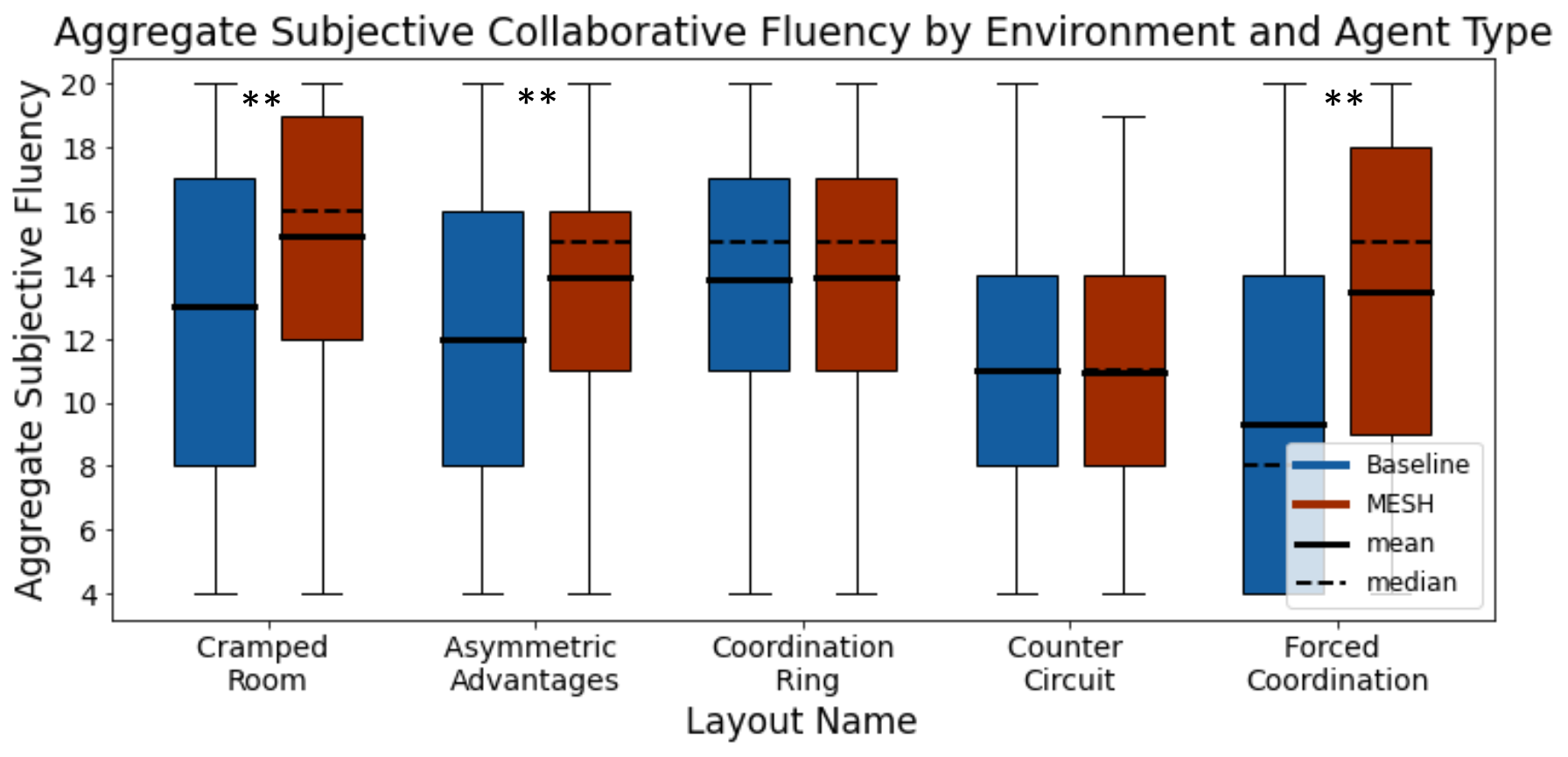}
\caption{Subjective collaborative fluency aggregated over the four measures was higher with MESH in three environments, with comparable numbers in the other two environments.}
\label{subj}
\vspace{-0.25in}
\end{figure}

\textbf{H4} is supported in the majority of environments. In the Cramped Room, Asymmetric Advantages, and Forced Coordination environment, participants rated the collaborative fluency of the interaction significantly higher when paired with the MESH model over the Baseline model (Fig \ref{subj}). There was a significant main effect of agent type on the aggregate subjective fluency measure ($F(1,124)=23.693, p<0.001, \eta^2 = 0.160$), and a significant interaction between the layout and agent type ($F(4,496)=12.294, p<0.001, \eta^2 = 0.090$). Post hoc testing with Bonferroni correction showed that for Cramped Room ($p=0.001$), Asymmetric Advantages ($p=0.002$) and Forced Coordination ($p<0.001$), there was a significant increase in self-reported interaction fluency ratings with the MESH agent (Fig \ref{subj}).  

\section{Discussion}
\label{sec:discussion}
Our results show that an agent designed for coordination on strategy may often positively impact task performance and collaborative fluency of human-robot teams. The benefits to task performance and fluency are exhibited in some, but not all, environments. Human-MESH teams outperformed human-Baseline teams in two environments: Cramped Room and Forced Coordination. The the other three environments did not exhibit significant differences in task performance between agents. In the three environments without performance improvement, the MESH algorithm for learning strategy-specific policies likely resulted in behavior similar to the baseline approach. The IRL reward learning under the set of predefined features may not have resulted in significantly different learned rewards for each strategy, causing the agents trained using the learned reward features to display similar behavior to the aggregate human model baseline.

Participants did not show strong preferences for either of the two agents, despite achieving higher score with the MESH agent. This result suggests that the differences between the two agents may have been too subtle to perceive. Agent behavior under a given strategy may not be identifiable from all states, but rather a subset of particular states. The policies trained on learned reward weights may have led to behavior similar across strategies and to the baseline for many states, leading to agent behavior that was not easily distinguishable by participants.

While task efficiency is one measure of agent performance, collaborative fluency can provide insight into the ability of human-robot teams to coordinate their actions on shared tasks. The interaction with MESH saw higher concurrent activity between the agent and human partner (4 of 5 environments), as well as lower agent idle time (all environments). Participants teaming with the MESH agent also generally experienced lower human idle time; however, since there was no significant interaction between layout and agent type, we did not perform post-hoc analyses on which environments this particularly occurred in. These measures of objective collaborative fluency are related: human and robot idle time influence the amount of concurrent activity. Participants perceived higher collaborative fluency in their interactions with the MESH agent in a majority (3 out of 5) of environments. The other two environments, Coordination Ring and Counter Circuit, which did not report higher subjective fluency, were also among the environments where the MESH agent did not improve task performance. This further suggests that there was not significant difference in agent behavior between strategy-specific policies and the baseline policy. Our results demonstrate that having an agent coordinate on collaborative strategy with a new human partner can offer potential benefits in task performance and collaborative fluency, but the effectiveness varies across environments. The MESH framework for training coordination agents does not underperform compared to the baseline in any environments, suggesting that developing an agent capable of coordinating with different behavioral strategies in human data can assist in better leveraging limited demonstrations. 

A key assumption of MESH is that a team that adopts the same strategy will perform better than a team that doesn't. While our study demonstrates that strategy coordination benefits performance and team fluency, there do exist scenarios in which this is not true: it is possible that two different task strategies may be complimentary, and the best response of a robot partner is to employ a complimentary, not identical, approach. We ran a simulated experiment comparing the performance of our strategy-specific policies with each other, and found that there were cases where teams of matching strategies performed worse than non-matching teams, illustrating an example of when direct matching may not be ideal. We additionally assume that the dataset of team observations contains a representative sample of varying team strategies from which we can learn a reasonable library of policies. Under this assumption, by strategically focusing agent training on a subset of the data, we can construct strategy-specific agent policies. 

This work represents a step towards developing collaborative agents capable of coordinating their task approaches with human partners. As data gathered about human behavior in different collaboration contexts must increase for better trained agents, this approach provides a candidate unsupervised approach for training agents with an understanding of human strategies and an ability to coordinate with learned, emergent strategies. In future work, we aim to consider alternative reward features, that may more strongly represent differences between strategies, and compare our method to planning-based approaches and approaches that infer the intentions of the human partner.

\section{Conclusion}
\label{sec:conclusion}

Coordination of joint actions between humans and robots is required for effective collaboration. We employ a notion of collaborative strategies as approaches teams take towards interweaving team members' actions in performing a shared task. In particular, agents must coordinate their strategies for approaching joint goals in order to facilitate fluent team collaboration. This ability to adapt necessitates the agent first identify existing collaborative strategies and learn policies that employ each strategy. In this work, we develop MESH: a computational framework for learning human collaborative strategies and training an agent with the ability to adapt to the strategies of unseen human partners. An unsupervised clustering-based approach is used for recognizing different teaming patterns from human data. We leverage the learned clusters to train a library of different policies that perform distinct collaborative strategies. We use the identified strategies to construct a Mixture-of-Experts model that adapts to unseen human partners. Results of our user study find that the MESH agent offers higher objective and subjective collaborative fluency over an existing baseline. We see this work as an investigation towards designing robot behavior that reasons about how different task-completion strategies interplay with those of other team members.

\vspace{-0.01in}
\section*{ACKNOWLEDGMENT}

This research is based upon work supported by the Defense Advanced Research Projects Agency, award number: \text{FP00002636}. This work was also supported by NSF IIS-2112633. We thank Timothy Hyun for assistance in data analysis.

\vspace{-0.2in}

\bibliographystyle{IEEEtran}
\bibliography{strategy}
\end{document}